\documentclass{article}
\usepackage[utf8]{inputenc}

\usepackage{xcolor}
\usepackage{hyperref}
\hypersetup{
    colorlinks=true,
    urlcolor=blue,
}
\usepackage{authblk}

\newcommand{\miles}[1]{\textcolor{green}{{\bf [MB]}}}
\renewcommand\Affilfont{\small} 

\makeatletter
\renewcommand\AB@affilsepx{\quad \protect\Affilfont}
\makeatother

\title{Understanding the Capabilities, Limitations, and Societal Impact of Large Language Models}
\author[1]{Alex Tamkin\thanks{Equal contribution}}
\author[2]{Miles Brundage$^{*}$}
\author[3]{\authorcr Jack Clark\thanks{Work carried out while employed at OpenAI}}
\author[1,3]{Deep Ganguli}
\affil[1]{Stanford University} \affil[2]{OpenAI} \affil[3]{AI Index}
\date{}

\begin{document}

\maketitle
\vspace{-.55in}
\section*{Introduction}

On October 14th, 2020, researchers from OpenAI, the Stanford Institute for Human-Centered Artificial Intelligence, and other universities convened to discuss open research questions surrounding \href{https://papers.nips.cc/paper/2020/hash/1457c0d6bfcb4967418bfb8ac142f64a-Abstract.html}{GPT-3}, the largest publicly-disclosed dense language model at the time. 

The meeting took place under Chatham House Rules. Discussants came from a variety of research backgrounds including computer science, linguistics, philosophy, political science, communications, cyber policy, and more. Broadly, the discussion centered around two main questions:
\begin{enumerate}
    \item \textbf{What are the technical capabilities and limitations of large language models?}
    The discussion touched on several key areas including: the surprising impact of scale on model capabilities, the difficulty in assessing whether large language models truly understand language, the importance of training models on multiple data modalities, and challenges in aligning model objectives with human values.
    
    \item \textbf{What are the societal effects of widespread use of large language models?}
    The discussion touched on several key areas including: difficulties in scoping all possible uses (or misuses) of general purpose language models, challenges organizations may face in model deployment, the potential for these models to algorithmically spread disinformation, difficulties in mitigating model bias (e.g., racial, gender, religious, etc.), and the impact of language model-based automation on the labor market. 

\end{enumerate}

While the conversation was collegial and productive, there was a sense of urgency to make progress sooner than later in answering these questions. Here, we provide a detailed summary of the discussion organized by the two themes above.\footnote{Since this is a summary of discussions, rather than a research paper, we do not include references. Rather, we hyperlink to relevant papers that were discussed at the workshop. For a more comprehensive set of references related to some of these issues, we point readers to the original \href{https://papers.nips.cc/paper/2020/hash/1457c0d6bfcb4967418bfb8ac142f64a-Abstract.html}{GPT-3 paper} and to recent work of \href{https://faculty.washington.edu/ebender/papers/Stochastic_Parrots.pdf}{Bender and Gebru et al} published a few months after this workshop.} We conclude with a list of potential future research directions inspired by the discussion.

\section{Technical Capabilities and Limitations}

\subsection*{Scale}
GPT-3 is one of the largest publicly-disclosed language models — it has 175 billion parameters and was trained on 570 gigabytes of text. For comparison, its predecessor, GPT-2 (which is functionally similar to GPT-3) has 1.5 billion parameters and was trained on 40 gigabytes of text. While GPT-2 displayed some zero-shot generalization to downstream tasks, GPT-3 further displayed the ability to learn more novel tasks when given examples in context. Participants found it remarkable that such capabilities emerge merely from scaling model and training data size.

One person remarked that the growth in model capabilities as they scale ``feels like a law of physics or thermodynamics" in its \href{https://arxiv.org/abs/2001.08361}{stability and predictability}. Several participants were optimistic that these trends would continue even for models much larger than GPT-3, yielding ever-stronger models capable of more advanced few-shot learning of new skills from a small number of training examples. 

One participant remarked that the scale of models like GPT-3 was reminiscent of large particle accelerator experiments, which require many people with diverse backgrounds to execute. For example, when training such large models, different teams with diverse expertise must collaborate to run experiments, build and maintain the computing infrastructure, develop the algorithms, and continuously interrogate the model’s capabilities for possible problems (e.g., bias, misuse, safety concerns, etc.). The latter point is referred to as ``red-teaming'' throughout the rest of this document.

\subsection*{Understanding}

What constitutes ``understanding'' in a language model, and does GPT-3 fulfill this definition? Some leaned towards definitions based on strong notions of intelligence, which require models to possess intentionality or the ability to respond to requests in the real world. Others suggested that there were even weaker notions of intelligence that models had yet to satisfy, including robustness to adversarial examples — data examples that easily confuse an AI system but not humans. Participants suggested that getting things ``mostly right" may not be sufficient for understanding if the model performs poorly on rare but important inputs. 

Another definition of understanding centered around the notion of causality, in that models that truly understand should grasp the causal relationship between features of the data and the desired behavior. Some argued that language models were destined to exploit “spurious correlations” or “shortcut features” inherent in the data, and thus lacked a true underlying causal model. However, one participant suggested a different view — that with enough data, language models could encounter “natural experiments” that could enable the model to learn causal relationships from observational data in a similar manner as human economists often do in their research. 

Some participants argued against binary thresholds for understanding, recalling that children and adults gradually acquire greater mastery over time. For example, one participant quoted a prominent physicist who quipped that he only understood thermodynamics the third time he taught it. Another participant pushed back against singular notions of understanding, noting debates between linguists and philosophers about whether meaning is derived from the relationship of expressions to each other or to some external ground truth.

Finally, some participants offered resistance to the focus on understanding, arguing that humans are able to accomplish many tasks with mediocre or even poor understanding, including a non-French speaker who recently won the French Scrabble championships. Some gently suggested that perhaps a judgment about whether GPT-3 understands language in the relevant way is irrelevant to successful performance of tasks. 

In a memorable line, one participant also remarked on the inverse problem of humans’ ability to understand large language models: ``GPT-3 is completely alien…it's the first thing I've seen where it's not a dumb thing to ask whether it's AGI.” Here, AGI refers to Artificial General Intelligence, or the ability of a machine to learn and understand anything a human can.

\subsection*{Multimodality}

Much of the conversation considered the importance of multimodal models — language models trained on text and data from other modalities, e.g., images, audio recordings, etc. Participants largely agreed in their predictions that large multimodal models will become more prevalent and enable more diverse capabilities.\footnote{In fact, shortly after the workshop, OpenAI released \href{https://openai.com/blog/dall-e/}{DALL-E}, which is a multimodal version of GPT-3 trained on both images and text.} However, some argued that GPT-3 is already trained on multimodal data, in that the training data contains prose, structured data tables, and computer code. Others suggested that the main benefit of multimodal training might be to improve the speed at which models acquire useful capabilities, as the interaction between different data modalities may provide a stronger learning signal than each data modality in isolation provides. Finally, some commented that no single additional modality was critical to language use, given that humans differ in the range of sensory modalities they have access to.

\subsection*{Alignment}

Participants discussed the need to better align model objectives with human values. For example, one participant mentioned some language models treat all symbols (e.g., nouns, prepositions, numbers, etc.) equally, but humans care much more about, for example, incorrectly stating someone's age than about misplacing a preposition. Several other participants emphasized the importance and challenge of better optimizing for factual accuracy and robustness to adversarial examples. Aligning human and model objectives was seen to be especially important for ``embodied" AI agents which learn through active interaction with their environment. Discussants emphasized the dual importance of developing better algorithms for ``steering" agents towards human values, as well as fostering cross-disciplinary collaborations to better clarify what ``human values" means, especially given diversity across individuals and communities and the prevalence of bias in available datasets.

\section{Effects of Widespread Use}

\subsection*{Capabilities}

GPT-3 has an unusually large set of capabilities, including text summarization, chatbot behavior, search, code generation, and essay generation. One discussant stated that such a large ``capability surface” makes it challenging to both scope the full array of uses (because GPT-3 can take in arbitrary inputs, it is \emph{a priori} impossible to anticipate all potential behaviors of the model) and to ensure their safety to people and societies. Participants noted that, by putting GPT-3 behind a controlled-access API, OpenAI is able to constrain the model’s use more easily than if they open sourced it. However, open questions remain. For example, who gets access and why? How can one provide model access to support a large community to red-team (interrogate the model for potential misuse and develop mitigation strategies) at scale?

\subsection*{Deployment}

Participants discussed several options for defining and addressing the ethical and societal challenges of deploying large language models. One suggestion was to increase the computing resources available to academia so that it would be easier for academics to do research that informs the deployment of large language models. Someone suggested that laws requiring disclosure of when AI is being used to generate text could be helpful in managing the effects of large language models. Another participant asked what metrics might be used to evaluate whether language models are having a societally beneficial effect, and there was general agreement that this is a challenging but important task. 

Several participants noted that OpenAI and other organizations will not have a monopoly on large language models forever. Participants suggested that developers may only have a six- to nine-month advantage until others can reproduce their results. It was widely agreed upon that those on the cutting edge should use their position on the frontier to responsibly set norms in the emerging field. Additionally, some participants pointed out that, due to standard advances in technology, it will only become easier for other actors to replicate models like GPT-3 over time. This further suggests the urgency of using the current time window, during which few actors possess very large language models, to develop appropriate norms and principles for others to follow.

\subsection*{Disinformation}

A major discussion point considered the deliberate misuse of language models for purposes such as generating disinformation. More specifically, models like GPT-3 can be used to create false, misleading, or propagandistic essays, tweets, and news stories \emph{de novo}. One participant was skeptical about the magnitude of these likely risks since many previous technologies (e.g. photography and Photoshop) sparked similar concerns and have already raised societal awareness of the risks of disinformation. Furthermore, while automated generation of disinformation may be feasible in principle, human labor may still be more cost-effective for such purposes. Others disagreed, and saw automated generation as much more cost-effective than training and paying humans to generate disinformation. Participants agreed that empirically investigating the economics of automated vs human generated disinformation is important. 

Thinking ahead, someone suggested considering a future in which language models can generate text that is not just coherent on commonly discussed topics, but highly persuasive on arbitrary topics. Another participant suggested that GPT-3 or other future language models could make disinformation hard or impossible to detect at the level of content, forcing reliance on metadata by online platforms. Relatedly, someone suggested that the existence of systems like GPT-3 should spur more use of cryptography to authenticate media.

\subsection*{Bias}

GPT-3 exhibits several racial, gender, and religious biases. One discussant analogized the difficulty of addressing language model bias to the problem of content moderation on online platforms — despite the difficult normative issues in both cases, there are still some areas of relative consensus and opportunities for mitigation. For example, online platforms agree on the need to address child pornography or egregious threats of violence, and the concept of “protected classes” in discrimination law provides a useful initial framework for thinking about some language model biases.

Several workshop participants noted that it is difficult to define what it means to mitigate bias in large language models in a universal manner, since appropriate language use is highly contextual. One participant noted that all datasets are biased in some ways, so the challenge is not eliminating all bias but addressing harmful biases according to some set of normative and/or legal criteria. Some suggested that companies like OpenAI do not have the appropriate standing and should not aim to make such decisions on behalf of society. Someone else observed that it is especially difficult to think about mitigating bias for multi-purpose systems like GPT-3 via changes to their training data, since bias is typically analyzed in the context of a particular use cases. 

Participants discussed a wide variety of possible means of addressing harmful biases in language models, including:

\begin{itemize}
    \item Changes to the initial training data to mitigate bias \emph{a priori}
    \item Training a separate model to filter content generated by a language model
    \item Fine-tuning a large language model on data with desired properties
    \item Tagging  data so that the model learns to distinguish among certain forms of content (see e.g. \href{https://arxiv.org/abs/1909.05858}{CTRL})
    \item Training models to be more \href{https://arxiv.org/abs/1906.07241}{“fact-aware”}
    \item Reinforcement learning with \href{https://openai.com/blog/learning-to-summarize-with-human-feedback/}{human feedback}
    \item Leveraging the model’s own knowledge to improve outputs (e.g., with careful prompt design)
    \item Developing more expansive suites of ``bias tests'' that models can be run through prior to deployment 
    \item Red-teaming the model at scale by engaging trusted partners to work with the model and through limited commercial offerings.
\end{itemize}

None of these approaches was considered a panacea. For example, steering a model with human feedback still raises the question of who the human labelers are or how they should be chosen, and content filters can sometimes undermine the agency of the very groups that they are intended to protect (e.g., marginalized groups reclaiming words or phrases that are used as slurs by majority groups). One participant argued that keeping a human in the loop of text generation is critical for addressing these issues. Some participants emphasized that certain use cases should be avoided given the limitations of existing techniques, and that text generation applications vary widely in terms of open-endedness and risk. For example, detecting regular expressions is much more tractable to do safely than managing a suicide hotline.

\subsection*{Economy}

Another theme of the discussion considered the economic implications of models like GPT-3. Participants observed that current jobs that involve reading or analyzing text vary widely in their desirability, with some being more enjoyable (e.g., creative writing or reading and summarizing reports) and others often being traumatizing or alienating (e.g., content moderation). This raises the question of when jobs, or what kinds of jobs, should or shouldn’t be automated by large language models. One participant suggested that leaving such decisions up to companies would likely have adverse consequences. Education was also mentioned as a societal area likely to be affected by large language models, via changes to the essay writing process as well as evaluation of text. One participant pointed out that providing API access to a variety of groups from different sectors of society can help provide an early signal of potential societal changes.

\section{Future Research Directions}

The following research questions were inspired by the discussion:

\begin{itemize}
    \item  Can we better understand why language models improve so much with scale? Can this enable us to build models which scale more efficiently? 
    \item What are the limits of scaling? Will scale lead to strong causal reasoning, symbolic manipulation, commonsense understanding, and robustness to a wider class of inputs? Or will different techniques be necessary? 
    \item How can we understand the limits of what large language models are capable of? Can we enable models to ask for help or clarification, or abstain when they are unsure? 
    \item How can we develop new neural network architectures and algorithms that enable efficient learning from diverse, multimodal data beyond text?
    \item What are the opportunities and tradeoffs involved in different approaches to steering the outputs of large-scale language models to be more aligned with human values? 
    \item How should access to models like GPT-3 be allocated, balancing considerations like security, replicability, and fairness?  
    What kinds of tests do we need to develop in order to qualify language models like GPT-3 as being safe or unsafe for use in particular contexts? 
    \item What can academia do to best position itself to develop guardrails for the industrial development of such models - including advocating for sufficient funding to replicate the compute resources required to train them? 
    \item How can we best foster cross-disciplinary collaboration to \href{https://arxiv.org/abs/1803.09010}{understand} and manage the biases in large datasets and model representations of such datasets? 
    \item How can we best characterize the potential ``threat landscape" for such models; e.g., do we need to spend more time worrying about how models like this could be used by profit-driven actors to generate lots of low-grade spam, or should we be more worried about state-based actors using models to generate persuasive text for use in disinformation campaigns?
    \item How cost-effective and skill-intensive would it be for malicious actors to misuse language models for various purposes, compared to alternative methods of achieving the same goals?

\end{itemize}

\end{document}